\documentclass[letterpaper]{article} %
\usepackage[draft]{aaai2026}  %
\usepackage{times}  %
\usepackage{helvet}  %
\usepackage{courier}  %
\usepackage[hyphens]{url}  %
\usepackage{graphicx} %
\usepackage{natbib}  %
\usepackage{caption} %
\usepackage{algorithm}
\usepackage{algorithmic}

\usepackage{newfloat}
\usepackage{listings}
\DeclareCaptionStyle{ruled}{labelfont=normalfont,labelsep=colon,strut=off} %
\floatstyle{ruled}
\newfloat{listing}{tb}{lst}{}
\floatname{listing}{Listing}
\usepackage{booktabs}
\usepackage{amssymb}
\usepackage{amsmath}
\usepackage{xcolor}

\title{ShoulderShot: Generating Over-the-Shoulder Dialogue Videos}
\author{
    Yuang Zhang\textsuperscript{\rm 1},
    Junqi Cheng\textsuperscript{\rm 1},
    Haoyu Zhao\textsuperscript{\rm 1,2},
    Jiaxi Gu\textsuperscript{\rm 1}\thanks{Corresponding author: imjiaxi@gmail.com},\\
    Fangyuan Zou\textsuperscript{\rm 1},
    Zenghui Lu\textsuperscript{\rm 1},
    Peng Shu\textsuperscript{\rm 1}
}
\affiliations{
    \textsuperscript{\rm 1}Tencent
    \textsuperscript{\rm 2}Fudan University

}

\usepackage{bibentry}

\begin{document}

\maketitle

\begin{abstract}
Over-the-shoulder dialogue videos are essential in films, short dramas, and advertisements, providing visual variety and enhancing viewers' emotional connection. Despite their importance, such dialogue scenes remain largely underexplored in video generation research.
The main challenges include maintaining character consistency across different shots, creating a sense of spatial continuity, and generating long, multi-turn dialogues within limited computational budgets. Here, we present ShoulderShot, a framework that combines dual-shot generation with looping video, enabling extended dialogues while preserving character consistency. Our results demonstrate capabilities that surpass existing methods in terms of shot-reverse-shot layout, spatial continuity, and flexibility in dialogue length, thereby opening up new possibilities for practical dialogue video generation. Videos and comparisons are available at {https://shouldershot.github.io}.
\end{abstract}

\section{Introduction}

Over-the-shoulder (OTS) dialogue videos are a fundamental element in storytelling across films, short dramas, and advertisements. They provide visual variety, enhance emotional engagement, and help guide the viewer’s attention during conversations. Despite their importance, such scenes remain underexplored in video generation research, where most efforts focus on single-shot or single-character sequences with limited duration.

\begin{figure}[ht]
    \centering
    \includegraphics[width=\linewidth]{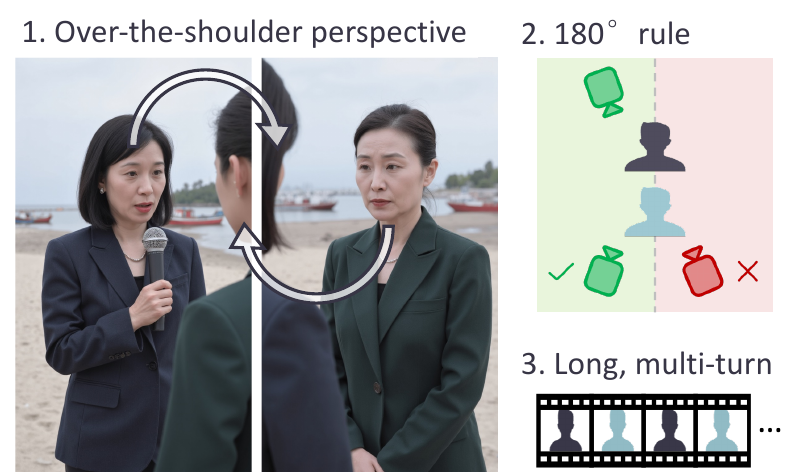}
    \caption{An example of over-the-shoulder (OTS) shots. Key aspects for visual continuity include: 1) OTS perspective with consistent character across shots; and 2) The 180$^\circ$ rule, keeping the camera on one side of the implied axis between subjects to prevent disorienting jumps. 3) These shots are often used for long, multi-turn dialogues.}
    \label{fig:task}
\end{figure}

As illustrated in Figure~\ref{fig:task}, achieving effective dual-shot over-the-shoulder dialogue videos requires addressing several key aspects. To ensure a seamless viewing experience, generating the \textit{over-the-shoulder perspective} is essential; the main subject in one shot appears as the same person whose shoulder is shown in the reverse shot when the camera switches.
To create a sense of spatial continuity and prevent confusion, adhering to the \textit{180$^\circ$ rule} is a standard practice; this guideline ensures the camera stays on one side of an imaginary line between characters, maintaining their consistent relative positions across shots~\cite{proferes2005film}.
Finally, to enable extended conversations, supporting \textit{long, multi-turn dialogues} within limited computational budgets is vital, necessitating efficient use of video generation models, given their typically high computational cost.

In this work, we ask: how can we generate dual-shot over-the-shoulder dialogue videos with consistent characters, adherence to the 180$^\circ$ rule, and longer dialogue durations? To address these challenges, we introduce ShoulderShot.

For character consistency, we adopt an in-context dual-shot generation method that ensures character consistency between shots by leveraging LoRA-finetuned text-to-image models~\cite{lhhuang2024iclora}. The model generates two consecutive shots side by side---in the left shot, the first person takes the main position, while the second person only shows their shoulder in the foreground; in the right shot, the second person takes the main position, while the first person only shows their shoulder in the foreground. Through training with a few samples, the model learns to maintain consistency between the main character in one shot and the corresponding shoulder figure in the other.

For adherence to the 180$^\circ$ rule, we enforce a fixed composition: the main character in the left shot always appears on the left side of the frame, and the main character in the right shot always appears on the right side. This ensures the characters' relative positions remain consistent across both shots, thereby enhancing audience understanding of the dialogue scene.

For longer dialogue durations, we adopt a looping video generation strategy. Specifically, we generate an 8-second looping video template for each shot. By randomly selecting the starting point of each loop, we introduce variations to the template each time it's used, reducing repetition and enhancing visual diversity. For even longer single shots, the template can be looped multiple times, maintaining coherence while extending the overall length.

Our contributions are threefold:
\begin{enumerate}
    \item We propose a dual-shot generation framework for over-the-shoulder dialogue scenes with character consistency and adherence to the 180$^\circ$ rule.
    \item We introduce a looping video generation strategy that enables the creation of extended, multi-turn dialogues under limited computational cost.
    \item Our approach produces high-quality dialogue videos that outperform existing methods in character consistency, shot layout, and video length.
\end{enumerate}

\section{Related Works}
Our work builds upon advancements in video generation, yet addresses distinct challenges in generating long over-the-shoulder dialogue videos. In this section, we review existing research in dialogue video generation, multi-shot character consistency, and long video generation.

\subsection{Dialogue Video Generation}
Recent advancements in video generation extend beyond single-speaker scenarios to dialogue scenarios. However, existing methods are limited in terms of over-the-shoulder shot composition and duration. MoCha~\cite{wei2025mocha} enables the generation of multi-character, turn-based over-the-shoulder shots, but its total length is limited to 5.3 seconds, which is insufficient for practical multi-turn dialogues.
InterActHuman~\cite{wang2025interacthuman}, on the other hand, supports longer video generation with character reference and region-specific audio conditioning. However, its shot layout often appears unnatural, as characters do not consistently face each other.
Complementing these, DualTalk~\cite{peng2025dualtalk} tackles dual-speaker 3D talking head 3D motion generation, but they do not consider over-the-shoulder positioning and character consistency. TV-Dialogue~\cite{wang2025tvdialogue} focuses on dialogue text generation based on video content and user-specified themes.

\subsection{Multi-Shot Character Consistency}
Keeping character consistency is vital for image and video generation models~\cite{ruiz2023dreambooth,chen2024anydoor,li2024photomaker,zhao2024magdiff,zhang2025mimicmotion}. It helps generate continuous and highly customized content, especially concerning human identity. In image generation models, the most common scenario involves synthesizing images that maintain a given character's appearance. DreamBooth~\cite{ruiz2023dreambooth} is an early work that fine-tunes a pre-trained text-to-image model, enabling it to bind a unique identifier to a specific subject. Following the same few-shot training paradigm, Photomaker~\cite{li2024photomaker} also effectively maintains character consistency across different scenes. Training-based methods~\cite{guo2024pulid,hu2024instruct} incur a significant training cost to extract features of different characters and inject them into the diffusion process for consistency maintenance. Some works~\cite{chen2024anydoor,tewel2024training} also try to use a zero-shot manner to achieve this task. Anydoor~\cite{chen2024anydoor} designs identity features to preserve appearance details while allowing versatile local variations.
Additionally, virtual try-on models~\cite{han2018viton,zhu2024m,he2022style,gou2023taming,morelli2023ladi,yang2024texture,morelli2022dress,li2023virtual}, as a specific application, also require high character consistency, particularly between the reference human and the clothing.

However, while the above methods achieve good performance, they fail to generate coherent content for multi-shot and multi-person scenarios. Similar but different from the methods~\cite{lhhuang2024iclora, sun2024generative}, our approach can generate multiple human images across different views, which is suitable for achieving over-the-shoulder dialogue video generation.

\subsection{Long Video Generation}
While diffusion-based video generation has advanced, generating long, coherent videos remains a challenge, typically limited to a few seconds.

One category of approaches focuses on generating long videos by seamlessly transitioning among consecutive video segments. Approaches like Gen-L-Video~\cite{wang2023gen} and MimicMotion~\cite{zhang2025mimicmotion} use latent fusion to ensure smoother transitions between segments. Autoregressive approaches~\cite{lvdm, voleti2022mcvd,chen2025skyreelsv2} generate indefinite videos by conditioning on previous frames, but they can suffer from quality degradation due to error accumulation. These methods, while extending video length, often incur significant computational costs that grow with the desired video length.

Alternatively, looping videos offer a path to extended durations. While looping videos can be achieved by switching the playback direction from forward to reverse, this approach often results in unnatural visuals, such as stiff motion and physically implausible effects. Mobius~\cite{bi2025mobius} proposes a latent shifting strategy that constructs a latent cycle and employs a latent shift mechanism to fully leverage the video diffusion model’s multi-frame latent denoising capabilities at each step, enabling the generation of seamlessly looping videos. However, this approach is primarily designed for looping short videos. While Mobius uses RoPE interpolation to extend the duration of single segments, both ROPE interpolation and extrapolation can lead to performance degradation and unstable behavior. Building on this, we introduce a strategy for generating long videos by looping video segments, where each segment can be as long as 10 seconds or more. This approach is better suited for over-the-shoulder dialogue videos.

\begin{figure*}[ht]
    \centering
    \includegraphics[width=\linewidth]{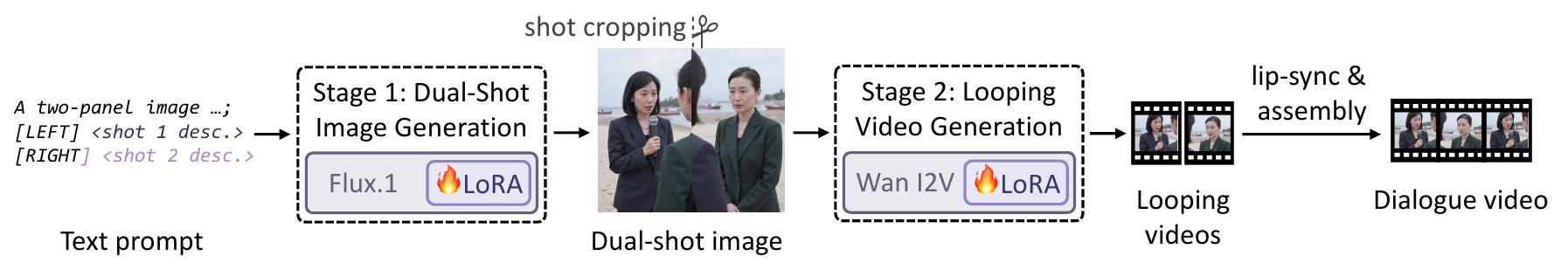}
    \vspace{-16px}
    \caption{Dual-shot dialogue video generation pipeline. In the image generation stage, a text prompt describing two shots guides a finetuned Flux model to generate a dual-shot image. Each cropped shot then independently generates a looping video with a finetuned Wan I2V model. Finally, these looping segments are lip-synced and assembled into the final dialogue video.}
    \label{fig:pipeline}
\end{figure*}

\section{Method}

In this section, we first present the overall pipeline, followed by detailed descriptions of dual-shot image generation, training data filtering, and looping video generation.

\subsection{Overall Pipeline}

The overall pipeline for generating dual-shot dialogue videos is illustrated in Figure \ref{fig:pipeline}. Our framework takes a dialogue script and scene description for two shots as input and proceeds through two main stages: (1) \emph{Dual-Shot Image Generation}, where a text description of the two shots is used to generate a dual-shot image using a LoRA-finetuned image generation model, ensuring character consistency and adherence to the 180$^\circ$ rule. (2) \emph{Looping Video Generation}, where each shot cropped from the dual-shot image is processed by a video generation model to create a looping video template. Finally, we extract segments from the looped video, synchronize the characters' mouth movements with the user-provided audio, and assemble the video clips into the final dialogue video.

\subsection{Dual-Shot Image Generation}
This stage generates a dual-shot image, serving as the visual reference for the dialogue video. The process uses a LoRA-finetuned text-to-image model.

Our approach is based on the observation that text-to-image models can generate coherent multi-panel images from a single prompt~\cite{lhhuang2024iclora}. Building on this capability, we leverage LoRA fine-tuning on a small set of high-quality data to produce two distinct shots within a single image---each represented as a panel, while maintaining character consistency across shots.

\paragraph{Prompt Formatting}
To guide the generation, we use a structured prompt template based on IC-LoRA~\cite{lhhuang2024iclora}: \texttt{A two-panel image split in the center; [LEFT] <description of the left shot> [RIGHT] <description of the right shot>.} This format provides separate descriptions for each shot, enabling precise control.

\begin{figure}[ht]
    \centering
    \includegraphics[width=\linewidth]{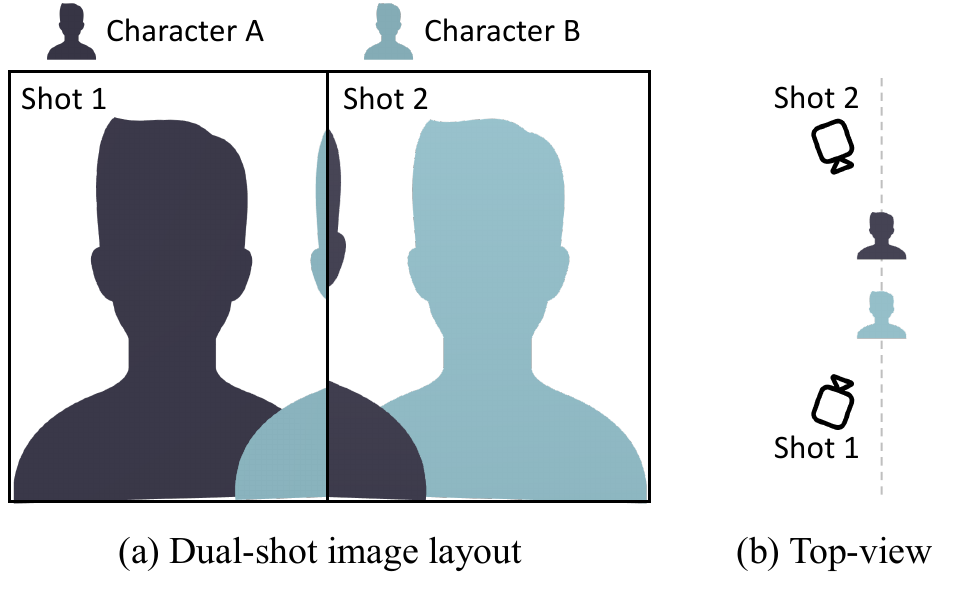}
    \vspace{-16px}
    \caption{Dual-shot image layout and camera positioning. (a) Dual-shot image layout, showing Character A consistently on the left and Character B on the right across both shots, which enforces the 180$^\circ$ rule. (b) Top-view showing the camera positions that maintain this spatial consistency.}
    \label{fig:shot}
\end{figure}

\paragraph{Training Data Composition}
Our training dataset consists of dual-shot image pairs, as shown in Figure~\ref{fig:shot}a. In each pair, the left shot features Character A as the main subject with Character B's shoulder in the foreground, while the right shot reverses this composition, featuring Character B as the main subject and Character A's shoulder in the foreground. This pairing enables the model to learn the spatial relationships of the characters in over-the-shoulder dialogues and their corresponding character consistency.

\paragraph{The 180$^\circ$ Rule}

The consistent spatial arrangement of characters within the generated dual-shot images, as depicted in Figure \ref{fig:shot}a, is foundational to our adherence to the 180$^\circ$ rule. This fixed compositional design directly addresses the rule by maintaining the relative positions of the characters across both shots. Specifically, Character A (the main subject in Shot 1) is always positioned on the left side of the frame, while Character B (the main subject in Shot 2) is consistently on the right side. This deliberate left-right placement, along with the complementary over-the-shoulder view of the off-screen character, ensures that the imaginary axis of action (as shown in Figure \ref{fig:shot}b) is maintained, thereby providing the audience with a clear spatial understanding throughout the dialogue.

\subsection{Training Data Filtering Pipeline}

\begin{figure}[t]
    \centering
    \includegraphics[width=\linewidth]{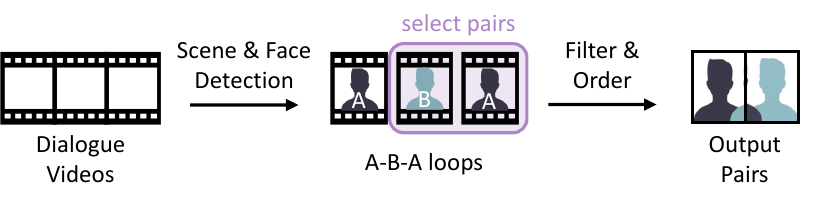}
    \caption{Training data filtering pipeline.}
    \label{fig:data}
\end{figure}

We design our training data filtering pipeline to mine high-quality samples of over-the-shoulder dialogue shots. Figure~\ref{fig:data} illustrates the overall process. We first collect a set of dialogue videos from the internet. Then, we perform \textit{scene detection} to segment the raw video footage into distinct scenes. To find dialogue shot pairs from abundant footage, we identify a common cinematic pattern in dialogue: the ``A-B-A loop.'' This refers to a sequence where a shot of character A is followed by a shot of character B, which then returns to a shot of character A, creating a repeating sequence of alternating perspectives. Following this, we perform \textit{face detection and recognition} to identify and track individual characters within these scenes. The pipeline then proceeds to \textit{find these A-B-A loops} within the detected scenes, from which the second B-A image pairs (the transition from B back to A, representing the reverse shot pair) are extracted. We then manually filter these extracted images, retaining only over-the-shoulder shots. Finally, a \textit{shot ordering} step is applied, ensuring that the shoulder appears on the right side of shot 1 and the left side of shot 2, adhering to our desired dual-shot image layout (Figure~\ref{fig:shot}) for maintaining spatial consistency.

\subsection{Looping Video Generation}

Since the speech content of a single character often has variable lengths, the length of the generated video of a single shot needs to be variable too. Commonly, video generation models have a maximum limit of generating video frames and suffer from high computation cost when generating a large number of frames. In this case, directly repeating a single clip of video in sequential or reverse order is a simple method for generating longer video shots. However, this will inevitably cause abrupt transitions between video clips. As a result, we design a sliding window diffusion strategy to generate seamless looping video. A looping video refers to a video that can be played infinitely in a loop starting from any point. In addition, to enrich the visual dynamics, generating more frames in a single loop is necessary. To this end, we integrate temporally overlapped diffusion to overcome the model's inherent frame count limitations.

Most Image-to-video models assume that the reference image serves as the initial frame of the video. This is often unfavorable for generating looping videos, because a looping video means that all generated video frames should be similarly close to the reference image, without any distinction. As a result, we finetune the base model of I2V with LoRA to unleash the ability of the model to generate videos without the assumption of treating the initial frame as a reference image. The training process can simply be done by modifying the data construction process. Concretely, during data construction, we use a random frame, instead of the initial one, from the input video as the reference image to guide the diffusion process.

\begin{figure}[ht]
    \centering
    \includegraphics[width=\linewidth]{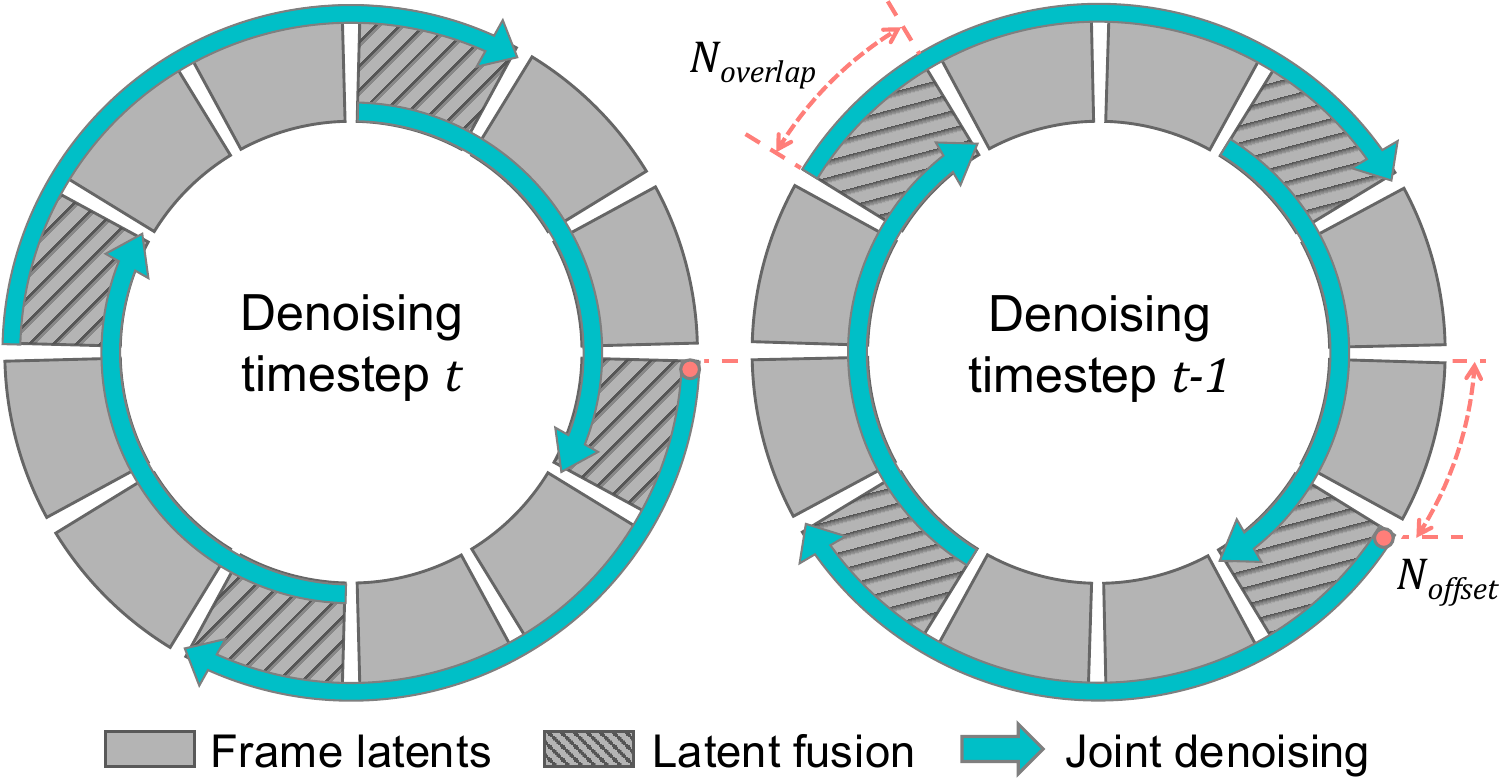}
    \caption{Frame latents are organized in a circular manner and partitioned into multiple overlapping segments. Each segment undergoes independent denoising by the I2V model, with latent fusion performed in the overlapping regions spanning $N_{\text{overlap}}$ latents. At the next time step, the segments are uniformly shifted by $N_{\text{offset}}$ latents, after which the model repeats the denoising and fusion processes on the newly arranged latents.}
    \label{fig:loop-video-diffusion}
\end{figure}

For generating looping videos with a long interval, we propose two strategies for model inference: sliding-window diffusion and temporally-overlapped diffusion. Figure~\ref{fig:loop-video-diffusion} shows a diagram of these strategies. Within each denoising step, all video frames are organized into a circular structure. In this circle, frames are segmented with a fixed overlapping so the denoising process can be applied on each segment consisting of a number of frames which can be handled by the I2V model. Each segment of frames is denoised individually, and a progressive fusion method is applied in the overlapped frames to generate the resulting latents for this timestep. The number of overlapped frames between adjacent segments is denoted as $N_{\text{overlap}},$ and the progressive fusion method is adopted from MimicMotion~\cite{zhang2025mimicmotion}. Taking inference efficiency into account, $N$ is set to the minimum value that guarantees no abrupt transition within the overlapped frames.

To make the resulting looping video seamless, the segment grouping of video frames needs to be variable so the influence of the frame position can be eliminated. To this end, we set a frame offset, denoted as $N_{\text{offset}}$, for the segment grouping. Every timestep during inference, we apply a frame offset to rearrange the frame segment grouping as Figure~\ref{fig:loop-video-diffusion} shows.

\begin{figure*}[t]
    \centering
    \includegraphics[width=\textwidth]{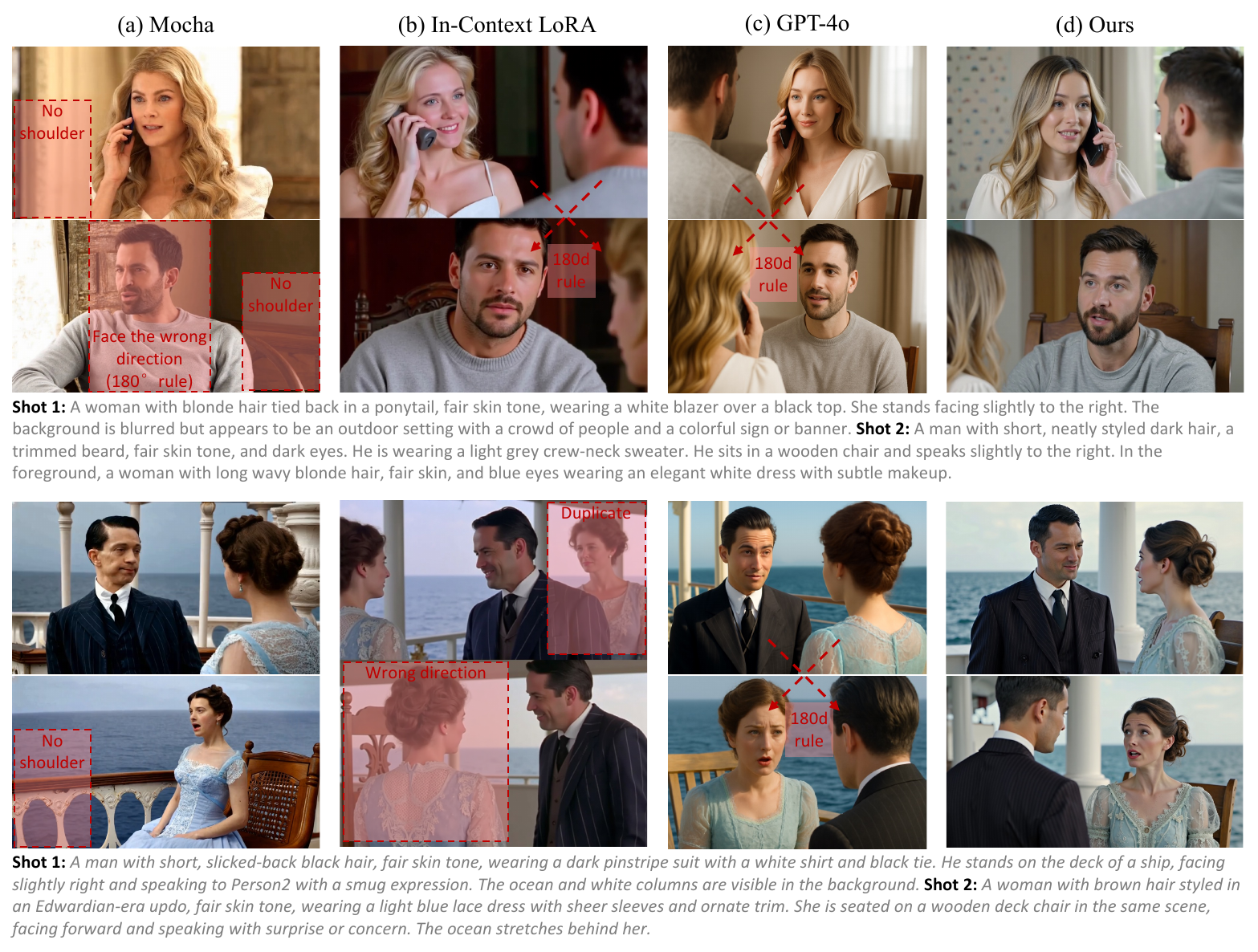}
    \vspace{-12px}
    \caption{Qualitative comparison of dual-shot over-the-shoulder images generated by existing methods and our results. (a) Mocha, (b) In-Context LoRA, (c) GPT-4o, (d) Ours. Our result shows better character consistency and shot layout.}
    \label{fig:qualitative}
\end{figure*}

\section{Experiments}

\subsection{Implementation Details}
For dual-shot image generation, our model is based on Flux-dev~\cite{flux2024}. We use LoRA~\cite{hu2022lora} for fine-tuning, configured with a rank of 16 and an alpha of 16. The model is trained for 5000 iterations with a learning rate of 2e-4 and a batch size of 4. We use the AI-Toolkit GitHub repository\footnote{https://github.com/ostris/ai-toolkit} for training. The data source for training contains 27 pairs of dual-shot images collected from online videos, specifically focused on over-the-shoulder dialogue scenes. We use Qwen-VL 2.5 to caption each single-shot image, then combine the captions of the two shots as the prompt for the dual-shot image.

For long video generation, we finetuned Wan 2.1 14B using LoRA (rank 16) with a learning rate of 2e-4 and a batch size of 16 for 1200 iterations on 469 videos. We use a fixed prompt for training. For loop video generation, we chose parameters of $N_{\text{overlap}}=4$ and $N_{\text{offset}}=9$.

\subsection{Qualitative Results}

We qualitatively evaluate ShoulderShot's performance by visually comparing its generated dual-shot over-the-shoulder images and videos against those produced by prominent existing methods.

\begin{figure}[t]
    \centering
    \includegraphics[width=\linewidth]{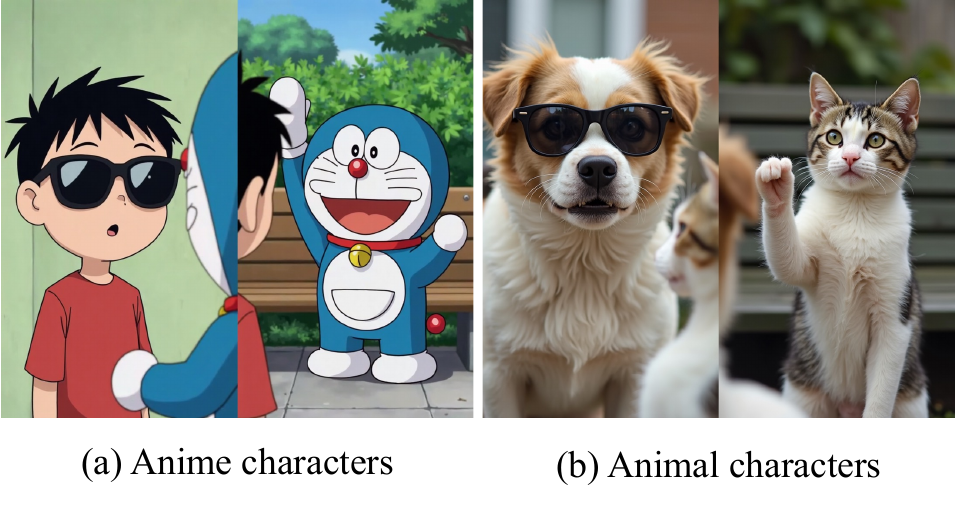}
    \vspace{-16px}
    \caption{Generalization of dual-shot over-the-shoulder images generation on anime and animal characters. The model maintains the desired shot layout across different styles.}
    \label{fig:generalization}
\end{figure}

First, we provide a qualitative comparison of generated dual-shot over-the-shoulder images. Figure \ref{fig:qualitative} showcases the output of ShoulderShot alongside results from MoCha~\cite{wei2025mocha}, in-context LoRA~\cite{lhhuang2024iclora} and GPT-4o~\cite{4oimage}. As demonstrated, ShoulderShot consistently exhibits superior character consistency and more accurate shot alignment, which are crucial for maintaining narrative coherence in dialogue scenes.

Next, we assess ShoulderShot's generalization capabilities across diverse character styles. Figure~\ref{fig:generalization} presents examples of dual-shot over-the-shoulder images featuring anime and animal characters. The results indicate that the model maintains character consistency and desired shot layout when applied to these different styles.

\begin{figure}[ht]
    \centering
    \includegraphics[width=\linewidth]{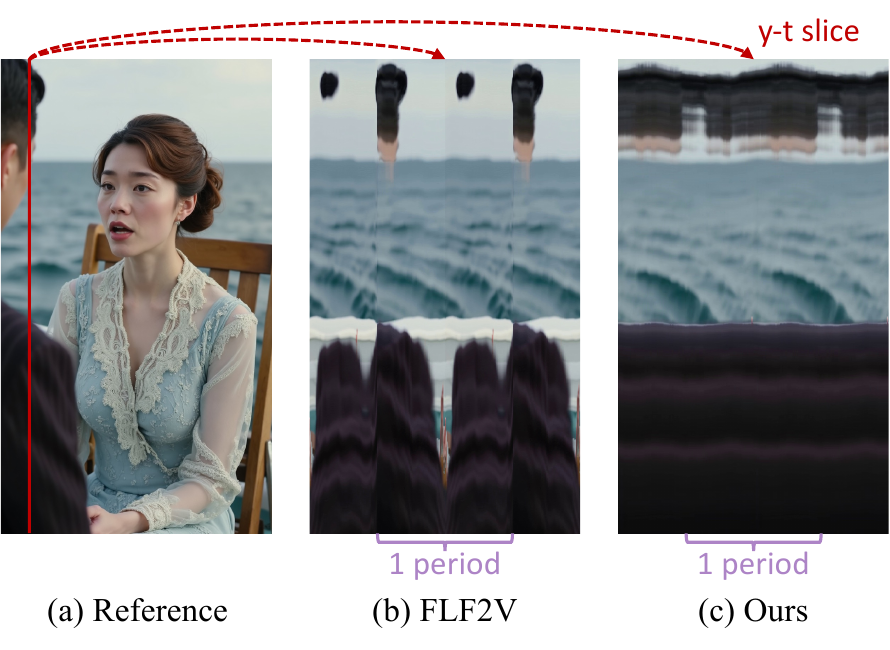}
    \vspace{-16px}
    \caption{Comparison of y-t slices for looping video. Y-t slice by FLF2V shows vertical stripe artifacts (indicating clip color abrupt change) and abrupt changes on the human face and ocean waves, indicating abrupt motion changes. Y-t slice by our method shows a smoother appearance, indicating seamless looping.}
    \label{fig:loop}
\end{figure}

Finally, we show ShoulderShot's ability to generate dual-shot over-the-shoulder videos. Figure~\ref{fig:loop} presents a comparison of y-t slices for looping video generation. The y-t slice from First-Last-Frame-to-Video (FLF2V) playback shows vertical stripe artifacts and abrupt changes in visual content, specifically on the human face and ocean waves. The y-t slice from our method exhibits a smoother appearance.

\subsection{Quantitative Comparison of Dual-Shot Images}

We quantitatively assess ShoulderShot's dual-shot image generation performance using several objective metrics, focusing on key aspects of over-the-shoulder dialogue scenes.

\paragraph{Methods for Comparison}
We compared ShoulderShot against several existing methods. (i.) \textit{In-context LoRA}~\cite{lhhuang2024iclora}: This method leverages the in-context abilities of Diffusion Transformers (DiTs) by concatenating images, performing joint captioning, and applying task-specific LoRA tuning on small datasets. We compare our method with their film storyboard generation capability. (ii.) \textit{GPT-4o}~\cite{4oimage} and \textit{Gemini 2.5 Flash}~\cite{gemini}: For these large language models, we rewrote the prompts to explicitly guide the generation of two over-the-shoulder shot/reverse-shot images. (iii.) \textit{MoCha}~\cite{wei2025mocha}: This audio-guided video generation method has the native end-to-end dual-shot video generation capability. We directly evaluate the frames of the videos in their publicly released results.

\paragraph{Evaluation Metrics}
Due to the lack of a reliable automated evaluation protocol, these metrics are assessed manually on a per-sample basis. Our metrics are defined as follows:
(i.) \textit{Over-the-Shoulder Shot (OTS)}: This metric assesses whether the shot includes the back of the shoulder of the person the main character is talking to, and maintains character consistency across shots. A score of 1 is given if both characters in their respective shots satisfy this criterion, and 0.5 if only one character's shot satisfies it.
(ii.) \textit{180$^\circ$ Rule}: This rule evaluates character positioning. For over-the-shoulder shots, it checks if both characters maintain the same relative position (e.g., if a character is on the left in the first shot, they remain on the left in the second shot). For non-over-the-shoulder shots, it verifies that the two characters face opposite directions and are not directly facing the screen.
(iii.) \textit{Eye Contact}: This metric specifically applies to over-the-shoulder shots and determines if the character's eyes are looking towards the other person. A score of 1 is given if both split shots satisfy this condition, and 0.5 if only one split shot satisfies it.

\paragraph{Dataset}
We use all the two-person two-clip prompts from the MoCha public evaluation set as our test dataset. The total number of prompts is 15, and all generated images and raw evaluations are included in the supplementary material.

We present a quantitative comparison of these metrics against existing methods in Table~\ref{tab:objective_metrics}. The scores are averaged across all samples. Our method consistently maintains the desired over-the-shoulder shot composition and maintains the 180$^\circ$ rule for spatial continuity.
MoCha~\cite{wei2025mocha} can generate multi-character, multi-shot dialogues, but it has a low proportion of over-the-shoulder shots (0.13).
Additionally, the spatial relationship between the two split shots is often inconsistent, with fewer satisfying the 180$^\circ$ rule (0.4). There are also instances where characters speak in the wrong direction, leading to low eye contact (0.07).
The original In-Context LoRA~\cite{lhhuang2024iclora} also has a certain probability of generating over-the-shoulder shot-reverse-shots, but its stability is far from sufficient.
Gemini~\cite{gemini} struggles with generating reasonable character placement and spatial relationships, resulting in significantly worse scores.
GPT-4o~\cite{4oimage} performs well in stable generation of over-the-shoulder shots ($0.97$), but lags in adherence to the 180$^\circ$ rule ($0.80$).
For all methods, the proportion of generated eyes correctly looking at the speaking counterpart is low ($<$0.6), indicating that achieving accurate eye contact remains a significant challenge for current methods.

\begin{table}[ht]
    \centering
    \caption{Quantitative comparison of objective metrics: Over-the-Shoulder (OTS) with consistent character, 180$^\circ$ rule adherence, and Eye Contact (Eye Cont.). Higher is better.}
    \label{tab:objective_metrics}
    \small
    \begin{tabular}{lccc}
        \toprule
        \textbf{Method} & \textbf{OTS} & \textbf{180$^\circ$ Rule} & \textbf{Eye Cont.} \\
        \midrule
        MoCha & 0.13 & 0.40 & 0.07 \\
        In-context LoRA & 0.73 & 0.33 & 0.47 \\
        Gemini 2.5 Flash & 0.33 & 0.00 & 0.20 \\
        GPT-4o & 0.97 & 0.80 & 0.57 \\
        ShoulderShot (Ours) & \textbf{1.00} & \textbf{1.00} & \textbf{0.60} \\
        \bottomrule
    \end{tabular}
\end{table}

\paragraph{Impact of LoRA Fine-Tuning}

\begin{figure}[t]
    \centering
    \includegraphics[width=\linewidth]{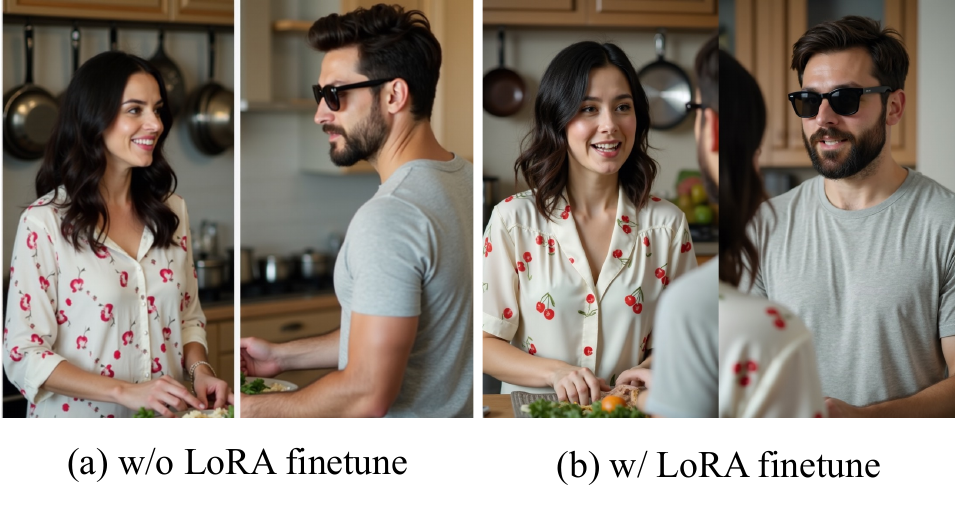}
    \vspace{-16px}
    \caption{Qualitative comparison of dual-shot image generation (a) without LoRA fine-tuning and (b) with LoRA fine-tuning. The fine-tuned model demonstrates improved adherence to the over-the-shoulder perspective and eye contact.}
    \label{fig:ft}
\end{figure}

We analyze the impact of LoRA fine-tuning on the Flux-dev base model for dual-shot image generation, focusing on both qualitative comparisons and objective metrics crucial for over-the-shoulder dialogue scenes. The evaluation assesses improvements in over-the-shoulder adherence, adherence to the 180$^\circ$ rule, and eye contact.

Qualitatively, Figure~\ref{fig:ft} compares images generated with the same prompt and seed, using both the base model and the LoRA fine-tuned model. Without LoRA fine-tuning, the base model's prior knowledge already enables the two individuals to face each other. This is consistent with the camera being positioned on the same side as both individuals, which helps maintain the 180$^\circ$ rule. Additionally, the base model is capable of generating two-panel images split in the center, albeit with some white margin. However, the base model struggles with generating the shoulder in the foreground.

Table~\ref{tab:lora_ablation} presents the results of this ablation study.
Before fine-tuning, the base model exhibits poor performance in generating clear over-the-shoulder shots (OTS: 0.00), and therefore cannot establish eye contact (Eye Cont.: 0.00). Interestingly, the base model demonstrates certain degree of inherent ability to maintain the 180$^\circ$ rule (0.87), as seen by the fact that the two people in the panels face each other (Figure~\ref{fig:ft}a). This makes LoRA training for over-the-shoulder shots adhering to the 180$^\circ$ rule feasible.

\begin{table}[ht]
    \centering
    \caption{Impact of LoRA fine-tuning in Over-the-Shoulder composition, 180$^\circ$ rule adherence, and Eye Contact (Eye Cont.). Higher is better.}
    \label{tab:lora_ablation}
    \small
    \begin{tabular}{cccc}
        \toprule
        \textbf{LoRA Finetune} & \textbf{OTS} & \textbf{180$^\circ$ Rule} & \textbf{Eye Cont.} \\
        \midrule
        $\times$ & 0.00 & 0.87 & 0.00 \\
        $\checkmark$ & 1.00 & 1.00 & 0.60 \\
        \bottomrule
    \end{tabular}
\end{table}

\subsection{Looping Video Generation}

To quantitatively evaluate our proposed looping video generation strategy, we compare it against two baseline methods: direct reverse playback and loop video generation using first and last frame control via Wan 2.1 FLF2V~\cite{wan2025}. For this comparison, we select a set of 20 video examples from the PATS dataset~\cite{ahuja2020style}.

We used two metrics for evaluation: FID-VID~\cite{balaji2019conditional} and FVD~\cite{unterthiner2018towards}. These metrics measure the distribution closeness to real videos, favoring perceptual quality and temporal consistency of generated videos at the grade of real videos, with lower values indicating better quality.

\begin{table}[h]
    \centering
    \caption{Quantitative comparison of loop video generation strategies: reversed vs. first-Last frame (FLF2V) playback and loop denoising (ours).}
    \label{tab:looping_ablation}
    \small
    \begin{tabular}{lcc}
        \toprule
        \textbf{Loop Video Generation Strategy} & \textbf{FVD ($\downarrow$)} & \textbf{FID-VID ($\downarrow$)} \\
        \midrule
        I2V with reverse playback & 368 & 3.12 \\
        I2V with FLF2V playback & 378 & 3.42 \\
        Loop denoising (ours) & \textbf{284} & \textbf{2.35} \\
        \bottomrule
    \end{tabular}
\end{table}

As shown in Table \ref{tab:looping_ablation}, our loop denoising strategy outperforms the baseline methods. It achieves an FVD of 284 and an FID-VID of 2.35. In contrast, direct reverse playback yields an FVD of 368 and an FID-VID of 3.12, while the FLF2V reversal method results in an FVD of 378 and an FID-VID of 3.42. These results demonstrate that our looping strategy produces videos with superior temporal coherence and perceptual quality, effectively reducing the unnaturalness often seen in baseline looping methods at segment boundaries, making it more suitable for extended dialogue durations.

\subsection{Limitations}
Despite its advancements, ShoulderShot has certain limitations. Firstly, achieving precise eye contact remains a challenge. This may be due to the pixel variations involved in controlling gaze direction, which are subtle but to which human observers are highly sensitive. Fine-tuning with LoRA might not adequately capture it, and this issue is not exclusive to our model, as other advanced models like GPT-4o also struggle with accurate eye contact despite consistently generating over-the-shoulder compositions.

Secondly, the current framework does not condition gesture generation on the specific dialogue content. This can lead to overly general or boring gestures, potentially degrading the naturalness and expressiveness of the generated dialogue videos. Furthermore, while lip-syncing is performed, its robustness with out-of-distribution (OOD) samples, such as animal videos, may be limited, potentially leading to noticeable synchronization failures in such scenarios. Nevertheless, our framework can readily integrate audio-conditioned talking video generation methods to mitigate these issues.

\section{Conclusion}

This work introduces ShoulderShot, a framework designed to generate over-the-shoulder dialogue videos. By combining a dual-shot generation method with a looping video strategy, ShoulderShot enables the creation of extended, multi-turn dialogues while adhering to practical computational constraints. Our experiments show its capabilities in generating shot-reverse-shot layouts that maintain character consistency and the 180$^\circ$ rule, offering more efficient and cohesive storytelling in video generation.

\bibliography{aaai2026}

\end{document}